%% file: bare_conf.tex
\documentclass[conference]{IEEEtran}

\usepackage{graphicx}
\usepackage{color}

\begin{document}

\title{Understanding the Impact of Precision Quantization on the Accuracy and
Energy of Neural Networks}%An Analysis of Accuracy-Power Trade-offs using Data Precision Scaling in Neural Networks}

% author names and affiliations
% use a multiple column layout for up to three different
% affiliations
\author{
\IEEEauthorblockN{Soheil Hashemi, Nicholas Anthony, Hokchhay Tann, R. Iris Bahar, Sherief Reda}
\IEEEauthorblockA{School of Engineering\\
Brown University\\
Providence, Rhode Island 02912\\
Email: \normalsize \{soheil\_hashemi, nicholas\_anthony, hokchhay\_tann, iris\_bahar, sherief\_reda\}@brown.edu}
% \and
% \IEEEauthorblockN{Soheil Hashemi}
% \IEEEauthorblockA{School of Engineering\\
% Brown University\\
% Providence, Rhode Island 02912\\
% Email: soheil\_hashemi@brown.edu}
% \and
% \IEEEauthorblockN{Soheil Hashemi}
% \IEEEauthorblockA{School of Engineering\\
% Brown University\\
% Providence, Rhode Island 02912\\
% Email: soheil\_hashemi@brown.edu}
% \and
% \IEEEauthorblockN{Soheil Hashemi}
% \IEEEauthorblockA{School of Engineering\\
% Brown University\\
% Providence, Rhode Island 02912\\
% Email: soheil\_hashemi@brown.edu}
% \and
% \IEEEauthorblockN{Soheil Hashemi}
% \IEEEauthorblockA{School of Engineering\\
% Brown University\\
% Providence, Rhode Island 02912\\
% Email: soheil\_hashemi@brown.edu}
}

% make the title area
\maketitle

% As a general rule, do not put math, special symbols or citations
% in the abstract
\begin{abstract}
Deep neural networks are gaining in popularity as they are used to generate state-of-the-art results for a variety of computer vision and machine learning applications. At the same time, these networks have grown in depth and complexity in order to solve harder problems. Given the limitations in power budgets dedicated to these networks, the importance of low-power, low-memory solutions has been stressed in recent years. While a large number of dedicated hardware using different precisions has recently been proposed, there exists no comprehensive study of different bit precisions and arithmetic in both inputs and network parameters. In this work, we address this issue and perform a study of different bit-precisions in neural networks (from floating-point to fixed-point, powers of two, and binary). In our evaluation, we consider and analyze the effect of precision scaling on both network accuracy and hardware metrics including memory footprint, power and energy consumption, and design area. 
We also investigate training-time methodologies to compensate for the reduction in accuracy due to limited bit precision and demonstrate that in most cases, precision scaling can deliver significant benefits in design metrics at the cost of very modest decreases in network accuracy. In addition, we propose that a small portion of the benefits achieved when using lower precisions can be forfeited to increase the network size and therefore the accuracy.
We evaluate our experiments, using three well-recognized networks and datasets to show its generality.
We investigate the trade-offs and highlight the benefits of using lower precisions in terms of energy and memory footprint.
\end{abstract}

\IEEEpeerreviewmaketitle

\section{Introduction}
\label{sec:Introduction}
\input{introduction.tex}

\section{Background}
\label{sec:Background}
\vspace*{-1mm}
\input{background.tex}

\section{Previous Work}
\label{sec:previous_work}
\vspace*{-1mm}
\input{previous_work.tex}

\section{Methodology}
\label{sec:methodology}
\vspace*{-1mm}
\input{methodology.tex}

\section{Experimental Results}
\label{sec:results}
\vspace*{-1mm}
\input{results.tex}

\section{Conclusion}
\label{sec:conclusion}
\vspace*{-1mm}
In this work, we perform an analysis of numerical precisions and quantizations in neural networks. We evaluate a broad range of numerical approximations in terms of accuracy, as well as design metrics such as area, power consumption, and energy requirements. %For our evaluations, we adopt a custom hardware accelerator capable of seamless processing of data, as the data is loaded into the buffers in parallel to the computation. 
We study floating-point arithmetic, different precisions of fixed-point arithmetic, quantizations of weights to be of powers of two, and finally binary nets where the weights are limited to one bit values. %Our results give a definitive answer to many of the open problems concerning the resistance of neural networks to low precision/quantization and offer optimal designs, maintaining the best accuracy and power trade-offs.
We also show that lower-precision, larger networks can be utilized which outperform the smaller full-precision counterparts in both energy and accuracy.
%For future work, we plan to evaluate multiple network architectures, which maybe more suited for each precision scheme, and larger benchmark such as ImageNet.
For future work, we plan on analytically investigating the correlations between network and datasets and their behavior in lower precision thereby effectively predicting the lower precision accuracy and hardware metrics. Further, we plan to develop architectures which support multiple radix point locations between layers. As discussed in \ref{sec:exp_2}, this feature may reduce the accuracy degradation significantly for lower precision networks.

\section*{Acknowledgment}
This work is supported by NSF grant 1420864 and by NVIDIA Corporation for their generous GPU donation. We also thank Professor Pedro Felzenszwalb for his helpful inputs.

\footnotesize
\bibliographystyle{abbrv}
%{\scriptsize \bibliography{refbib}}
{\scriptsize \bibliography{refbib}}

% \begin{thebibliography}{1}
% \bibitem{IEEEhowto:kopka}
% H.~Kopka and P.~W. Daly, \emph{A Guide to \LaTeX}, 3rd~ed.\hskip 1em plus
%   0.5em minus 0.4em\relax Harlow, England: Addison-Wesley, 1999.
% \end{thebibliography}

\end{document}

%% file: introduction.tex
%!TEX root =  bare_conf.tex

In the recent years, deep neural networks (DNN) have provided state-of-the-art results in many different applications specifically related to computer vision and machine learning. %The utilization of such networks in many real world applications highlights the importance of efficient and low-power implementations, capable of processing data with high throughput. %While solutions based on general purpose CPUs and GPUs are currently feasible, the explosive increase in power and computational demand of these networks has led to design of many different hardware ASIC accelerators. 
One dominant feature of neural networks is their high demand in terms of memory and computational power thereby limiting solutions based on these networks to high power GPUs and data centers. In addition, such high demands have led to the investigation of low power ASIC accelerators where designers are free to assign dedicated resources to increase the throughput. However, memory accesses and data transfer overheads play an important part in the total computation time and energy. When using accelerators, as a solution to data transfer overheads, specialized buffers have been introduced, thereby isolating the data transfer from the computation and enabling the memory subsystem to load the new data while the computation core is processing the previously loaded data.

Neural networks show inherent resilience to small and insignificant errors within their calculations. This error tolerance originates from the inherent tolerance of the applications themselves, and the training nature of the networks, where some of the errors are compensated by relearning and fine tuning the parameters. In this light, techniques proposed by approximate computing, such as approximate arithmetic, are an attractive option to lower the power consumption and design complexity in neural networks accelerators. However, as demonstrated by Chen {\it et al.}~\cite{diannao} and Tann {\it et al.}~\cite{Tann16}, the dominant portion of power and energy of hardware neural network accelerators is consumed in the memory subsystem, limiting the scope of arithmetic approximation. In this light, one particularly effective solution is reducing the bit-width required to represent the data.

While many accelerators have been proposed using different bit-precisions, most of these studies have been ad-hoc and give little to no explanation for choosing the specific precision. In particular, an evaluation of different precisions on the performance of the networks, considering both hardware metrics and inference accuracy, is not available. Such a study would provide researchers with better guidance as to the trade-offs available by such networks. In this paper we aim to address this issue by providing a quantitative analysis of different precisions and available trade-offs. More specifically, our paper makes the following contributions:

\begin{itemize}
\item We perform a detailed evaluation of a broad range of networks precisions, from binary weights to single precision floating-points, as well as several points in between.

\item We utilize learning techniques to improve the lost accuracy by taking advantage of the training process to increase the accuracy.

\item We evaluate our designs for both accuracy and hardware specific metrics, such as design area, power consumption, and delay, and demonstrate the results on a Pareto Frontier, enabling better evaluation of the available trade-offs.

%\item We evaluate the performance of different precision using two well-known network architectures, LeNet and ALEXNet, while classifying images from MNIST and CIFAR-10 datasets to demonstrate practical application of these trade-offs.

%\item We propose increase in depth of the network in order to overcome the accuracy degradation. We demonstrate low-precision designs capable of achieving equivalent accuracy performance compared to smaller floating-point network while offering significant benefits in power, area, and memory requirements.

\item Exploiting the benefits of lower precisions, we propose increasing the network size to compensate for accuracy degradation. Our results showcase low precision networks capable of achieving equivalent accuracy compared to smaller floating-point networks while offering significant improvements in energy consumption and design area.

\end{itemize}

The rest of the paper is organized as follows. In Section~\ref{sec:Background}, we briefly summarize the basics of neural networks, and in Section~\ref{sec:previous_work} we review related work. Then, in Section~\ref{sec:methodology} we describe various precisions and network training techniques used in our evaluations and argue  for increasing network size to recoup accuracy loss in lower precision networks. The results from our evaluations are provided in Section~\ref{sec:results} and  in Section~\ref{sec:conclusion} we summarize our finding and provide suggestions for future work.

%% file: background.tex
%!TEX root =  bare_conf.tex

\begin{figure}[t]
%\vspace{-0.2in}
   \begin{center}
   
    \includegraphics[scale=0.265]{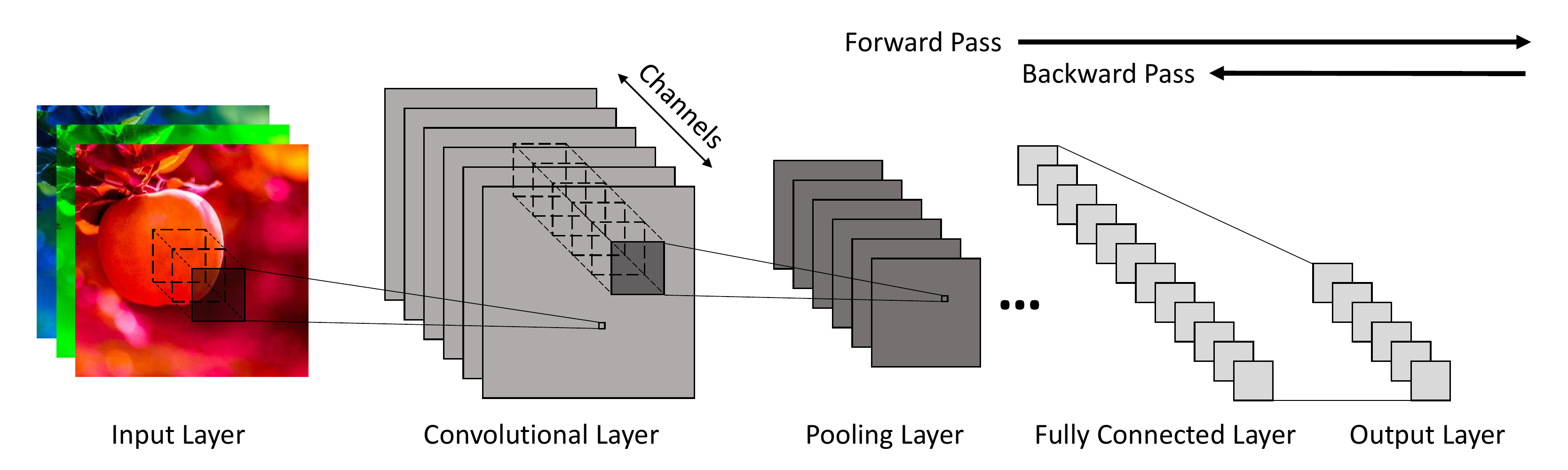}
    \vspace{-0.1in}
    \caption{The structure of a typical deep neural networks.}
    \label{fig:net}
    \end{center}
    \vspace{-0.3in}
\end{figure}

% \begin{figure}[t]
% %\vspace{-0.2in}
%    \begin{center}
%     \includegraphics[scale=0.30]{Images/neuron.pdf}
%     \vspace{-0.1in}
%     \caption{The structure of a neuron.}
%     \label{fig:neuron}
%     \end{center}
%     \vspace{-0.3in}
% \end{figure}

Deep neural networks are organized in layers where each layer is only connected to the layers immediately before and after it. Each layer gets its input from the previous layer and feeds it to the next layer after some layer-specific processing. Figure~\ref{fig:net} shows the general structure and connectivity of the layers. As show in the figure, each layer consists of several channels. Deep Neural networks, in general, consist of a combination of three main layer types: convolutional layers, pooling layers, and fully connected layers.

In typical neural networks the dominant portion of the computation is performed in the convolution layers and fully connected layers, while pooling layers simply down-sample the data. More specifically, channels in convolutional and fully connected layers are comprised of neuron units where each neuron performs a weighted sum of its inputs before feeding the result to a nonlinearity function.
The intermediate values between layers are called feature maps, as they each abstract some structure in the input image. From a data perspective, neural networks operate on two main set of parameters: input data and intermediate feature maps, and network parameters (or weights). Since inputs and feature maps are treated similarly by the network, similar precisions are used for their representation. However, 
%as will be shown in the next section, 
numerical precision of the network parameters can be changed independently of the input precision.

While the input data is assumed to be given for each network, the flexibility of neural networks arises from their ability to adapt their response to a specific input by training the network parameters. More specifically, use of neural networks comprises two phases, a training process during which the network parameters are learned, and a test phase which performs the inference and classification of the test data. In the training phase, neural networks usually utilize a backpropagation algorithm during which the classification error is propagated backwards using partial gradients. Network parameters are then updated using stochastic gradient descent. After training and in the test phase, the learned network is utilized in the forward phase to classify the test data. As discussed later, the main complexity of using lower precision in these networks arises due to the learning process.

%% file: previous_work.tex
%!TEX root =  bare_conf.tex

The high demand of DNNs, in terms of complexity and energy consumption, has shifted  attention to low-power accelerators. Many works have proposed implementing neural networks on FPGAs~\cite{Gokhale,Farabet:2009}, or as an ASIC accelerator~\cite{Kim,Temam}. In all these works, different precisions have been utilized with little or no justification for the chosen bit-width.

Chen {\it et al.} proposed Eyeriss, a spatial architecture along with a dataflow aimed at minimizing the movement energy overhead using data reuse~\cite{eyeriss}. For their implementation, a 16-bit fixed-point precision is utilized. 
Sankaradas {\it et al.} empirically determine an acceptable precision for their application~\cite{Sankaradas:2009} and reduce the precision to 16-bit fixed-point for inputs and intermediate values while maintaining 20-bit precision for weights. A FPGA-based accelerator is proposed by Zhang {\it et al.}, where single precision floating-point arithmetic has been utilized~\cite{Zhang:2015}. While this work offers a brief comparison between resources required for floating-point and fixed-point arithmetic logic in FPGAs, no discussion of accuracy is provided. Chakradhar {\it et al.} propose a configurable co-processor where input and output values are represented using 16 bits while intermediate values use 48 bits~\cite{Chakradhar:2010}. 
%A high-throughput accelerator using 16-bit fixed-point arithmetic is proposes by Chen {\it et al.}~\cite{Chen:2014}. In this work a comparison of 16-bit fixed-point precision is provided against 32-bit floating point precision. However, no comprehensive comparison for different precisions or different components is provided.

%On the other hand, 
Many works have successfully integrated techniques commonly used in approximate computing to lower the computation and energy demands of neural networks. A feedforward neural network is proposed by Kung {\it et al.}, where approximations are introduced to lower-impact synapses~\cite{Kung}.
%Here, approximations are added to the network through utilization of approximate multipliers or injecting zeros to the lower bits of a value therefore reducing the effective precision~\cite{Kung}.
Venkataramani {\it et al.} propose an approximate design where error-resilient neurons are replaced with lower-precision neurons and an incremental training process is used to compensate for some of the added error~\cite{Venkataramani:2014}. However, no specifications for the bit precision range used in the experiments are provided. Tann {\it et al.} propose an incremental training process during which most of the network can be turned off to save power~\cite{Tann16}. The neurons are then turned on during run-time if deemed necessary for correct classification. In this work, 32-bit floating-point representation was used.

While use of limited precision in neural networks has been proposed before~\cite{Lin,Courbariaux,Rastegari}, there exists no comprehensive exploration of their effect on energy consumption and computation time in reference to network accuracy. A recent publication by Gysel {\it et al.} provides an analysis of precision on network accuracy; however, the design parameters are not evaluated~\cite{Gysel}. Our objective is to precisely quantify the effect of each numerical precision or quantization on all aspects of the networks focusing specially on hardware metrics.

%% file: methodology.tex
%!TEX root =  bare_conf.tex

Here, first in Section~\ref{sec:meth_1}, we discuss the range of precisions and quantizations considered in our evaluation. We also briefly discuss the network training techniques used to minimize the accuracy degradation due to the limited precision. Finally, in Section~\ref{sec:meth_2} we propose two expanded network architectures to compensate for the accuracy drop.

\subsection{Evaluated Precisions and Train-Time Techniques}
\label{sec:meth_1}
We consider a broad range of numerical precisions and quantizations,  
%to conduct a comprehensive study. This spectrum consists of 
from 32-bit floating-point arithmetic to binary nets, as well as several precision points in between.  We summarize them below:

\subsubsection{Floating-Point Arithmetic}
This is the most commonly used precision as it generates the state-of-the-art results in accuracy. However, floating-point arithmetic requires complicated circuitry for the computational logic such as adders and multipliers as well as large bit-width, necessitating ample memory usage. As a result, this precision is 
%usually deployed in designs with few constraints on power, and therefore is 
not suitable for low-power and embedded devices.

\subsubsection{Fixed-Point Arithmetic}
Fix-point arithmetic is less computationally demanding as it simplifies the logic by fixing the location of the radix point. This arithmetic also provides the flexibility of a wide range of accuracy-power trade-offs by changing the number of bits used in the representation. 
In this work, we evaluate 4-, 8-, 16- and 32-bit precisions. To improve accuracy, we allow a different radix point location between data and parameters~\cite{Gysel}. However, we refrain from evaluating bit precisions that are not powers of 2 since they result in inefficient memory usage that might nullify the benefits.
%While these results can easily be interpolated based on the provided data points, 
\subsubsection{Power-of-Two Quantization}
Multipliers are the most demanding computational unit for neural networks. As proposed by Lin~\cite{Lin}, limiting the weights to be in the form of $2^i$, enables the network to replace expensive, frequent, and power-hungry multiplications with much smaller and less complex shifts.
%, thereby reducing  power and computation demands in the process. 
In our evaluations, we consider power of two quantization of the weights while representing the inputs with 16-bit fixed-point arithmetic.

\subsubsection{Binary Representation}
Recent work suggests that neural networks can generate acceptable results using just 1-bit weight representation~\cite{CourbariauxBD15}. While work by Courbariaux suggests binarizing activation between network layers, it does not binarize the input layer~\cite{Courbariaux}. For this reason, our accelerator would still need to support multi-bit inputs. Thus, we %perform all of our experiments using 16-bit fixed-point data and 1-bit weight.
evaluate the binary net using one bit for weights, while using 16-bit fixed-point representation for the inputs and feature maps. 

\begin{figure}[t]
%\vspace{-0.2in}
   \begin{center}
    \includegraphics[scale=0.33]{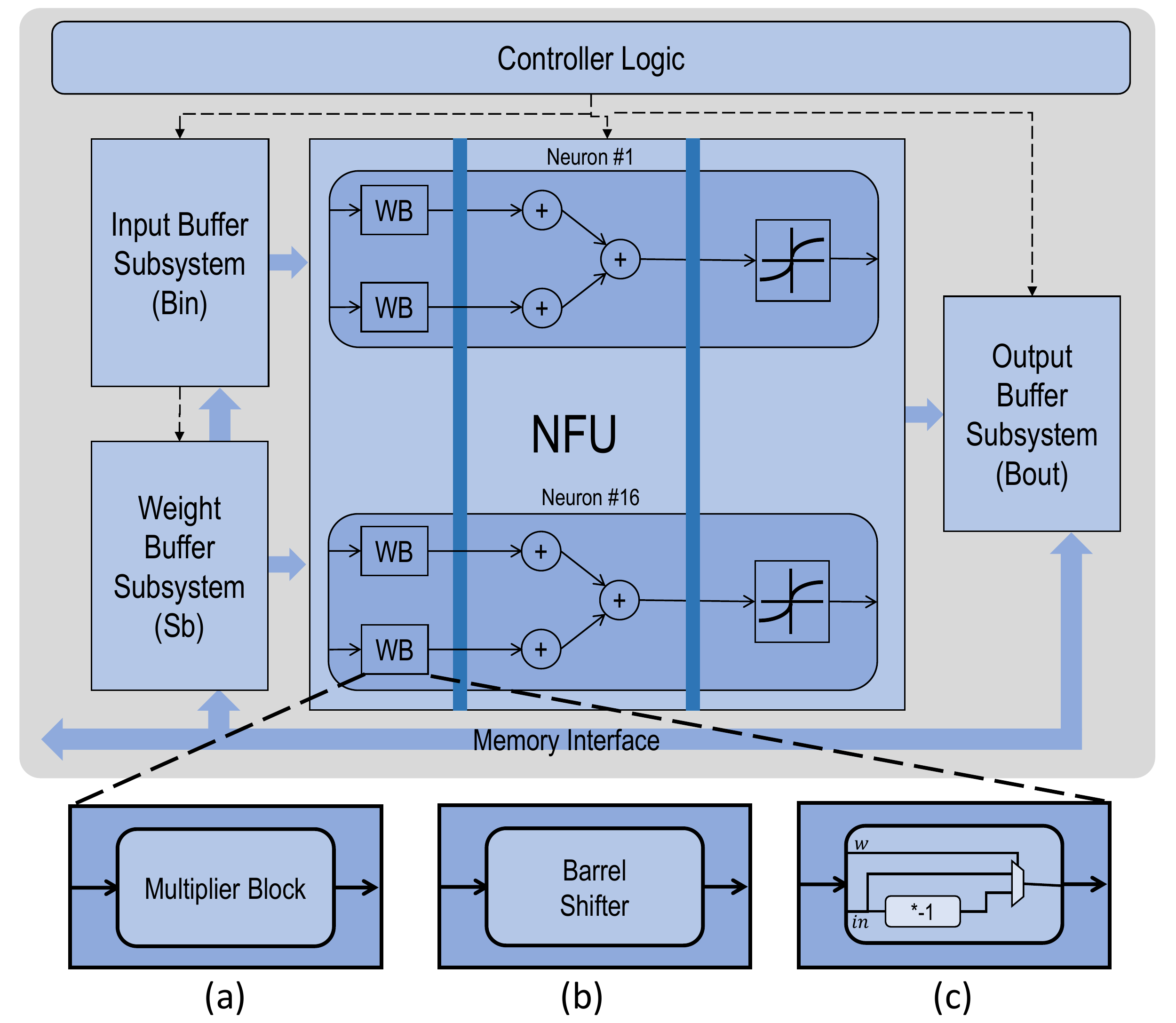}
    \vspace{-0.1in}
    \caption{The hardware model used for our experiments. The first stage (WB) has different variants for (a) floating-point and fixed-point arithmetic, (b) powers of two quantization, and (c) binary network.}
    \label{fig:hw}
    \end{center}
    \vspace{-0.3in}
\end{figure}

\vspace{0.05in} {\it Hardware Accelerator:} For our experiments, we adopt a tile-based  hardware accelerator similar to DianNao~\cite{diannao}. We implement 16 neuron processing units each with 16 synapses. Figure~\ref{fig:hw} shows our hardware implementation. 
%used for our tests. 
As illustrated in the figure, three separate memory subsystems are used to store the intermediate values and outputs and buffer the inputs and weights. These subsystems are comprised of an SRAM buffer array, a DMA, and control logic responsible for ensuring that the data is loaded into the buffers and made available to the neural functional unit (NFU) 
at the appropriate clock cycle without additional latency.
%can process the data without any concerns for data movement. 
The NFU pipelines the computation into three stages, weight blocks (WB), adder tree, and non-linearity function. As shown in Figure~\ref{fig:hw}, the weight blocks will be modified to accommodate for different precisions and quantizations as needed. In the case of binary precision, 
%as the first stage is reduced significantly, 
we merge the first two pipeline stages, effectively leading to a two stage NFU, in order to reduce the runtime. Furthermore, the size of all buffers and the control logic are modified according to the precision.

{\it Training Time Techniques:} We include a training phase in our experiments to enable the network to determine appropriate weights and adapt to the lower precision. 
%In the next subsection, we provide details on techniques utilized in the training algorithm to improve the accuracy when reduced due to lower precision.
%To obtain a deployment ready neural network, the network requires a training phase, during which network parameters, e.g. weights and biases, are learned.
Training processes, in nature, require high precision in order to converge to a good minima as the increments made to the parameters can be extremely small. On the other hand, if the network is made aware of its inference restrictions (in our case, the limited precision), the training process can potentially compensate for some of the errors by fine-tuning the parameters and therefore improve the accuracy at no extra cost. 

While the effects of reduced precision are analytically complicated to formulate as part of the training process~\cite{Gupta}, intuitive techniques can be utilized to improve the test phase accuracy.
One approach proposed in~\cite{Tann16} is to utilize a set of full precision weights, trained independently, as the starting point of a re-training process, in which the weights and inputs are restricted to the specified precision.
%To be more specific, the networks are first trained using full precision and then these learned parameters are used to initialize the limited precision parameters during a precision-aware re-training. In the re-training process, weights and inputs are represented with the specified limited precision of those discussed in Section~\ref{sec:meth_1}.
This approach assumes that by using lower precisions, close to optimal performance can be obtained if a local search is performed around the optimal set of parameters as learned with full precision.

A second approach for improving the accuracy is to utilize weights with different precisions in different parts of the training process, as proposed by Courbariaux {\it et al.}~\cite{CourbariauxBD15}. They solve the zero-gradient issue by keeping two sets of weights: one in full precision and one in the selected lower precision. The network is then trained using the full precision values during backward propagation and parameter updates, while approximating and using low precision values for forward passes. This approach allows for the accumulation of small gradient updates to eventually cause incremental updates in the lower precision.

In our approach, we train all of the low precision networks using a combination of the first and second approaches. We initialize the parameters for lower precision training from the floating point counterpart. Once initialized, we train by keeping two sets of weights.

\subsection{Expanded Network Architectures}
\label{sec:meth_2}
While significant savings in power, area, and computation time can be achieved using lower precisions, even a small degradation in accuracy can prohibit their use in many applications. However, we observe that, due to the nature of neural networks, the benefits obtainable by using lower precisions are disproportionately larger than the resulting accuracy degradation. This opens a new and intriguing dimension, where the accuracy can be boosted by increasing the number of computations while still consuming less energy. We therefore propose increasing the number of operations by increasing network size, as needed to maintain accuracy while spending significantly less for each operation.

In this light, in Section~\ref{sec:results}, we showcase two significantly larger networks and demonstrate that even by significantly increasing the size of the network, low precision can still result in improvements in energy consumption while eliminating the accuracy degradation. We discuss the specifications of the two larger networks in Section~\ref{sec:results}.

% The motivation behind our custom network is that (as shown in Section~\ref{sec:exp_2}) while significant savings in design metrics can be achieved, in all designs there is a degradation in accuracy, albeit insignificant, prohibiting the use of these precisions in many applications. Our key observation is that, due to the nature of neural networks, the benefits obtainable are disproportionately larger that then accuracy degradation. This opens a new and intriguing dimension, where the accuracy can be boosted by increasing the number of computation while still consuming less energy. In other words, we are proposing to increase the number of operations as needed to maintain accuracy while spending significantly less for each operations.

% In this light, as the most intuitive solution, we propose to use a larger network while keeping the precision low. Here, we evaluate two significantly larger network to demonstrate that even by increasing the size of network by a significant order, low-precision can result in less energy consumption in comparison to a smaller network using full-precision.

%% file: results.tex
%!TEX root =  bare_conf.tex

\begin{table}[t]
  \footnotesize
  \caption{Benchmark Networks Architecture Descriptions.}
  \vspace{-0.05in}
  \centering
  \begin{tabular}{c|c|c}
    \hline
    MNIST & SVHN & CIFAR-10 \\ %\hline
    LeNet \cite{lenet} &  ConvNet \cite{convnet} & ALEX \cite{cifar10} \\
    \hline
    28$\times$28$\times$1 & 32$\times$32$\times$3 & 32$\times$32$\times$3 \\
    conv 5$\times$5$\times$20 & conv 5$\times$5$\times$16 & conv 5$\times$5$\times$32 \\
    maxpool 2$\times$2 & maxpool 2$\times$2 & maxpool 3$\times$3 \\
    conv 5$\times$5$\times$50 & conv 7$\times$7$\times$512 & conv 5$\times$5$\times$32 \\
    maxpool 2$\times$2 & maxpool 2$\times$2 & avgpool 3$\times$3 \\
    innerproduct 500 & innerproduct 20 & conv 5$\times$5$\times$64 \\
    innerproduct 10 & innerproduct 10 & avgpool 3$\times$3 \\
                    &                 & innerproduct 10 \\
    \hline
  \end{tabular}
  
  \label{table:networks}
\end{table}

\subsection{Experimental Setup}
\label{sec:exp_1}
%\vspace*{-2mm}
We evaluate our designs both in terms of accuracy and design metrics (i.e., power, energy, memory requirements, design area). To measure accuracy, we adopt Ristretto~\cite{Gysel}, a Caffe-based framework~\cite{Caffe} extended to simulate fixed-point operation. We modify Ristretto to accommodate our techniques, as needed. In different experiments, we ensure that all design parameters except for the bit precision are the same. This is critical to ensure the isolation of the effects of bit precision from any other factor.

We compile our designs using Synopsys Design Compiler using a 65 nm industry strength technology node library. We use a 250 MHz clock frequency and synthesize in nominal processing corner. We design our accelerator to have a zero timing slack for the full-precision accurate design. We confirm the functionality of our hardware implementation with extensive simulations. As before, we ensure that all other network parameters, including the frequency, are kept constant across different precision experiments.

{\it Benchmarks:}
We consider three well-recognized neural network architectures utilized with three different datasets, MNIST~\cite{mnist} using the LeNet~\cite{lenet} architecture, SVHN using CONVnet~\cite{convnet}, and CIFAR-10~\cite{cifar10} using the network described by Alex Krizhevsky~\cite{cifar10} (Here we refer to this network as ALEX). For all cases, we randomly select 10\% of each classification category from the original test set as our validation set. 
To showcase the benefits from increasing the network size while using lower precision, we evaluate two networks as summarized in Table \ref{table:networks2}. Here, we focus on CIFAR-10 since MNIST and SVHN do not provide a large range in accuracy differences between various precisions and quantizations. As summarized in Table~\ref{table:networks2}, we evaluate two larger variations of the ALEX network: (1) ALEX+, where the number of channels in each convolutional layer is doubled, and (2) ALEX++, where the number of channels is doubled when the feature size is halved~\cite{VGG}. As shown in Section~\ref{sec:exp_2}, this methodology results in significant improvements in accuracy while still delivering significant savings in energy.

\begin{table}[t]
  \footnotesize
  \caption{ALEX Larger Network Architecture Descriptions.}
  \vspace{-0.05in}
  \centering
  \begin{tabular}{c|c}
    \hline
    \multicolumn{2}{c}{CIFAR-10} \\ %\hline
    ALEX+ & ALEX++ \\
    \hline
    32$\times$32$\times$3 & 32$\times$32$\times$3 \\
    conv 5$\times$5$\times$64 & conv 3$\times$3$\times$64 \\
    maxpool 3$\times$3 & maxpool 2$\times$2 \\
    conv 5$\times$5$\times$64 & conv 3$\times$3$\times$128 \\
    avgpool 3$\times$3 & maxpool 2$\times$2 \\
    conv 5$\times$5$\times$128 & conv 3$\times$3$\times$256 \\
    avgpool 3$\times$3 & maxpool 2$\times$2 \\
    innerproduct 10 & innerproduct 512 \\
    				& innerproduct 10 \\
    \hline
  \end{tabular}
  
  \label{table:networks2}
\end{table}

\subsection{Results}
\label{sec:exp_2}
\begin{figure}[t]
%\vspace{-0.2in}
   \begin{center}
    \includegraphics[scale=0.34]{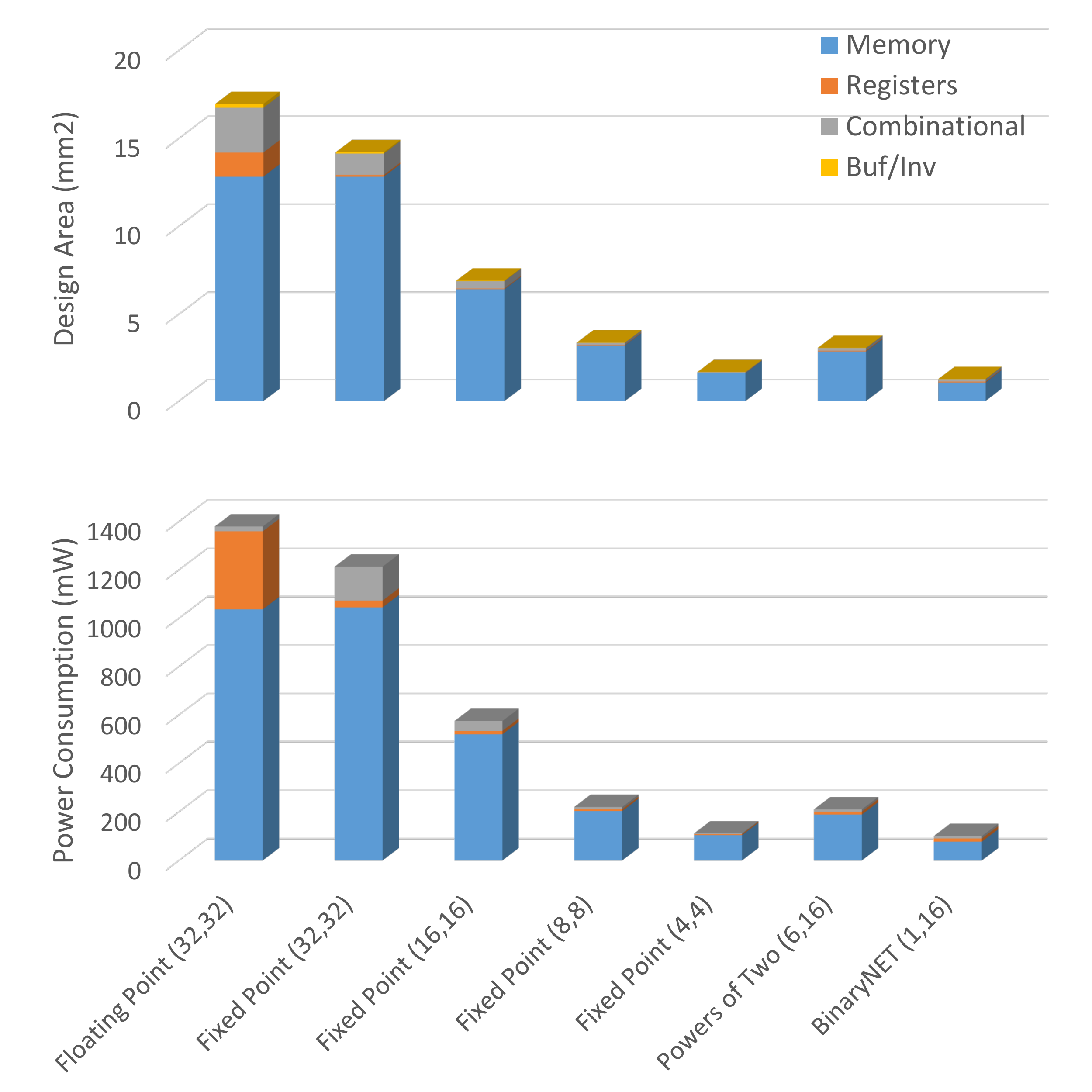}
    \vspace{-0.2in}
    \caption{The breakdown of design area and power consumption using different precisions.}
    \label{fig:power_break}
    \end{center}
    \vspace{-0.2in}
\end{figure}

%\vspace*{-2mm}
Figure~\ref{fig:power_break} shows the breakdown of power and area for the accelerator in the cases investigated. Values shown as ($w,in$) represent the number of bits required for representing weight and input values, respectively. Note, that these graphs do not reflect the power consumption of the main memory. As shown in the figure, the majority of the resources, both in power and design area, are utilized in the memory buffers necessary for seamless operation of the computational logic. To be more specific, in our experiments, the buffers consume between {75\%-93\%} of the total accelerator power, while using {76\%-96\%} of the total design area. These values highlight the necessity of approximation approaches targeting the memory footprint.

Table~\ref{table:power} summarizes the design metrics of the accelerator for each of the numerical precisions considered. In order to maintain a fair comparison, we keep all the other parameters, such as the frequency, number of hardware neurons, etc., constant among different precisions. Changing the frequency or the accelerator parameters (other than precision) adds another dimension to the design space exploration which is out of the scope of our work.

\begin{table}[t]
  \footnotesize
  \caption{Design metrics of the evaluated numerical precisions and quantizations.}
  \vspace{-0.05in}
  \centering
  \begin{tabular}{l|r|r|r|r}
    \hline
     	  & \multicolumn{1}{|l|}{Design} & \multicolumn{1}{|l|}{Power} & \multicolumn{1}{|l|}{Area} & \multicolumn{1}{|l}{Power} \\

		 & \multicolumn{1}{|l|}{Area} & \multicolumn{1}{|l|}{Cons.} & \multicolumn{1}{|l|}{Saving} & \multicolumn{1}{|l}{Saving} \\

Precision ($w,in$) & \multicolumn{1}{|l|}{($mm^2$)} & \multicolumn{1}{|l|}{($mW$)} & \multicolumn{1}{|l|}{($\%$)} & \multicolumn{1}{|l}{($\%$)} \\
    \hline
    Floating-Point (32,32) & 16.74 & 1379.60 & 0 & 0 \\
    Fixed-Point (32,32) & 14.13 & 1213.40 & 15.56 & 12.05 \\
    Fixed-Point (16,16) & 6.88 & 574.75 & 58.92 & 58.34 \\
    Fixed-Point (8,8) & 3.36 & 219.87 & 79.94 & 84.06 \\
    Fixed-Point (4,4) & 1.66 & 111.17 & 90.07 & 91.94 \\
    Powers of Two (6,16) & 3.05 & 209.91 & 81.78 & 84.78 \\
    Binary Net (1,16) & 1.21 & 95.36 & 92.73 & 93.08 \\
    \hline
  \end{tabular}
  \vspace{-0.2in}
  \label{table:power}
\end{table}

We evaluate the accuracy of the networks, as well as energy requirements for processing each image for each of our benchmarks. Table~\ref{table:acc_energy} summarizes the results for MNIST and SVHN datasets. We were able to achieve little to no accuracy drop for all but one of the network precisions in the MNIST classification. In the case of SVHN, however, while keeping the network architecture constant, the 4-bit fixed-point and binary representations failed to converge.
For SVHN dataset, for instance in the case of powers of two network, we are able to achieve more than 84\% energy saving with an accuracy drop of approximately 2\%. Note that as we keep the frequency constant the processing time per image changes very marginally among different precisions. Additional runtime savings can be achieved by increasing the frequency or changing the accelerator specification which is not explored in this work.

\begin{table}[t]
  \footnotesize
  \caption{The Accuracy, per image inference energy, and the energy savings achievable using each of the evaluated precisions. For each dataset, energy savings are in reference to the full-precision implementation.}
  \vspace{-0.05in}
  \centering
  \setlength\tabcolsep{2pt} % default value: 6pt
  \begin{tabular}{l|r|r|r|r|r|r}
    \hline
     	  & \multicolumn{3}{|c|}{MNIST} & \multicolumn{3}{|c}{SVHN} \\ \cline{2-7}
 & \multicolumn{1}{|l|}{Class.} & \multicolumn{1}{|l|}{Energy} & \multicolumn{1}{|l|}{Energy} & \multicolumn{1}{|l|}{Class.} & \multicolumn{1}{|l|}{Energy} & \multicolumn{1}{|l}{Energy} \\
Precision ($w,in$) & \multicolumn{1}{|l|}{Acc. ($\%$)} & \multicolumn{1}{|l|}{($uJ$)} & \multicolumn{1}{|l|}{Sav. ($\%$)} & \multicolumn{1}{|l|}{Acc. ($\%$)} & \multicolumn{1}{|l|}{($uJ$)} & \multicolumn{1}{|l}{Sav. ($\%$)} \\
    \hline
    Floating-Point (32,32) & 99.20 & 60.74 & 0 & 86.77 & 754.18 & 0 \\
    Fixed-Point (32,32) & 99.22 & 52.93 & 12.86 & 86.78 & 663.01 & 12.09 \\
    Fixed-Point (16,16) & 99.21 & 24.60 & 59.50 & 86.77 & 314.05 & 58.36 \\
    Fixed-Point (8,8) & 99.22 & 8.86 & 85.41 & 84.03 & 120.14 & 84.07 \\
    Fixed-Point (4,4) & 95.76 & 4.31 & 92.90 & NA & NA & NA \\
    Powers of Two (6,16) & 99.14 & 8.42 & 86.13 & 84.85 & 114.70 & 84.79 \\
    Binary Net (1,16) & 99.40 & 3.56 & 94.13 & 19.57 & 52.11 & 93.09 \\
    \hline
  \end{tabular}
  \vspace{-0.15in}
  \label{table:acc_energy}
\end{table}

The reduction in precision also reduced the required memory capacity for network parameters, as well as the input data. We quantify our memory requirements for all the network architectures using different bit precisions. In our experiments, for the full-precision design, network parameters require approximately 1650KB, and 2150KB, and 350KB of memory for LeNet, CONVnet, and ALEX, respectively. Since there is a direct correlation between bit precision and network memory requirements, the memory footprint of each network reduces from 2$\times$ to 32$\times$ for different bit precisions. Note, we do not utilize any of recent parameter encoding and compression techniques, and such techniques are orthogonal to our work.

As discussed in Section~\ref{sec:meth_2}, we propose that a portion of the benefits from using low precision arithmetic can be exploited to boost the accuracy to match that of the floating point network while spending some portion of the energy savings by increasing the network size. Here, we showcase the benefits from our proposed methodology on CIFAR-10 dataset. The summary of the performances for the ALEX as well as the two larger networks (ALEX+ and ALEX++) is provided in Table \ref{table:larger_network}. Here, we do not report the results for fixed-point (32,32) for ALEX+ and ALEX++ as its energy saving is not competitive compared to other precisions. Also, the fixed-point (4,4) fails to converge for all three networks on CIFAR-10 and the respective rows have been removed from the table. Furthermore, we find that the accuracy for fixed-point++ (8,8) is lower in comparison to the other networks with the same precision. We observe that for this network, there is a significant difference in the range of parameter and feature map values and as a result, 8 bits fails to capture the necessary range of the numbers.

As shown in the table, lower precision networks can outperform the baseline design in accuracy while still delivering savings in terms of energy. The parameter memory requirements for the full-precision networks are roughly 350KB, 1250KB, and 9400KB for ALEX, ALEX+, and ALEX++ respectively. As discussed previously, the memory footprint reduces linearly with parameter precision when reducing the precision.

\begin{table}[t]
  \footnotesize
  \caption{Network performance for different precision on CIFAR-10 dataset and using ALEX, ALEX+, and ALEX++. Energy savings are in reference to the ALEX full-precision implementation.}
  \vspace{-0.05in}
  \centering
  \begin{tabular}{l|r|r|r}
  \hline
 & \multicolumn{3}{|c}{CIFAR-10} \\ \cline{2-4}
 & \multicolumn{1}{|l|}{Class.} & \multicolumn{1}{|l|}{Energy} & \multicolumn{1}{|l}{Energy} \\
Precision ($w,in$) & \multicolumn{1}{|l|}{Acc. ($\%$)} & \multicolumn{1}{|l|}{($uJ$)} & \multicolumn{1}{|l}{Sav. ($\%$)} \\
    \hline
    Floating-Point (32,32) & 81.22 & 335.68 & 0 \\
    \hline
    Fixed-Point (32,32)    & 79.71 & 293.90 & 12.45 \\
    \hline
    Fixed-Point (16,16)    & 79.77 & 136.61 & 59.30 \\
    Fixed-Point+ (16,16)   & 81.86 &  491.32 & 1.5$\times$ More \\
    Fixed-Point++ (16,16)   & 82.26 &  628.17 & 1.9$\times$ More \\
    \hline
    Fixed-Point (8,8)      & 77.99 &  49.22 & 85.34 \\
    Fixed-Point+ (8,8)     & 78.71 &  177.02 & 47.27 \\
    Fixed-Point++ (8,8)     & 75.03 &  226.32 & 32.59 \\
    \hline
    Powers of Two (6,16)   & 77.03 &  46.77 & 86.07 \\
    Powers of Two+ (6,16)  & 77.34 &  168.21 & 49.89 \\
    Powers of Two++ (6,16)  & 81.26 &  215.05 & 35.93 \\
    \hline
    Binary Net (1,16)      & 74.84 &  19.79 & 94.10 \\
    Binary Net+ (1,16)     & 77.91 &   71.18 & 78.80 \\
    Binary Net++ (1,16)    & 80.52 &   91.00 & 72.89 \\
    \hline
  \end{tabular}
  \vspace{-0.15in}
  \label{table:larger_network}
\end{table}

\begin{figure}[t]
%\vspace{-0.2in}
   \begin{center}
    \includegraphics[scale=0.35]{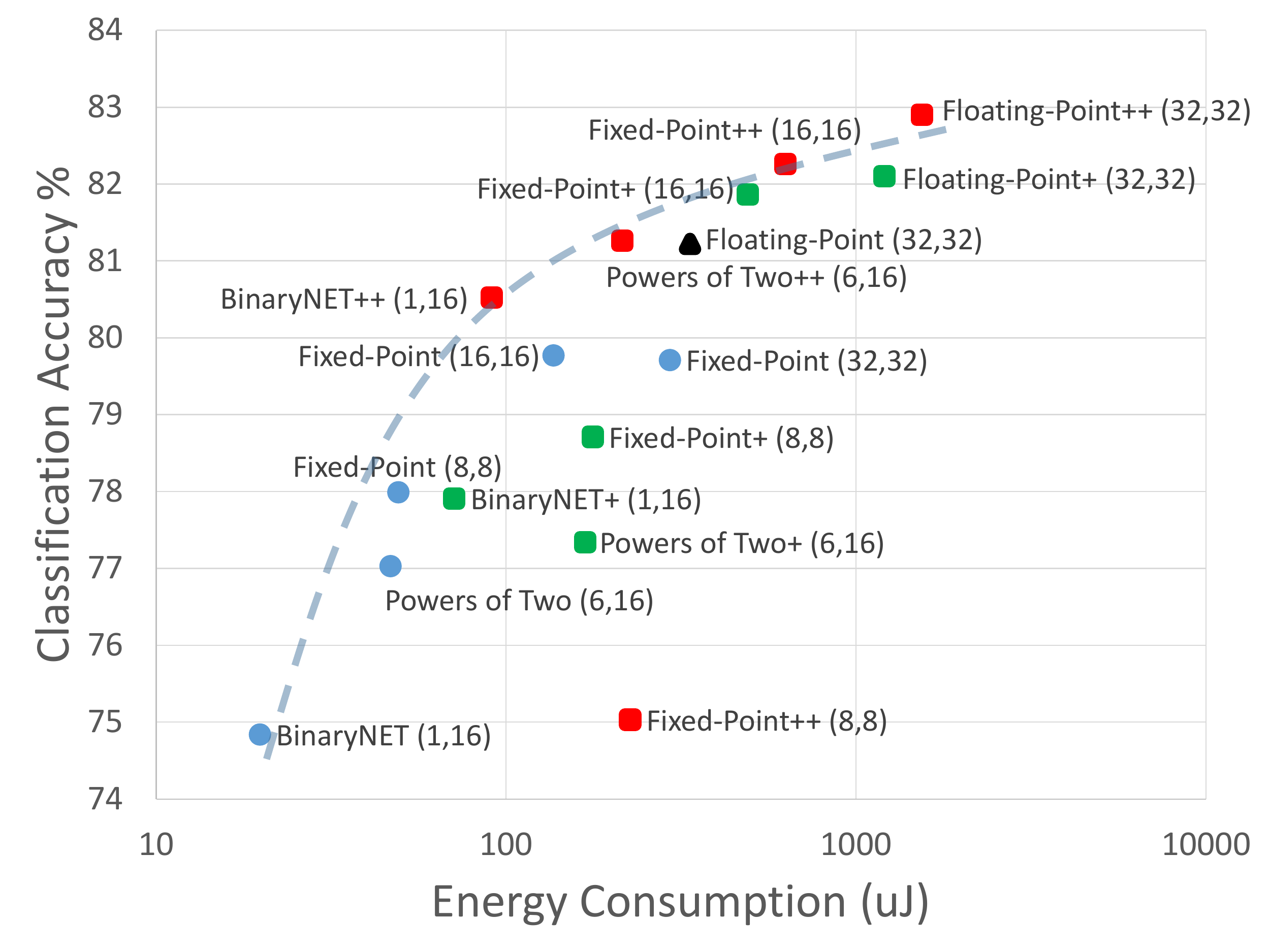}
    \vspace{-0.3in}
    \caption{The Pareto Frontier plot of the evaluated design point for CIFAR-10 testcase. The x axis is plotted in logarithmic scale to cover the energy range of all the designs. Here, the black point indicates the initial full-precision design, the blue points indicate the lower precision points, while the red and green points show the results from the larger networks.}
    \label{fig:pareto}
    \end{center}
    \vspace{-0.3in}
\end{figure}

The available trade-offs in terms of accuracy and energy using different precisions and expanded networks are plotted in Figure~\ref{fig:pareto} for the CIFAR-10 testbench. The figure highlights the previous argument that a wide range of power and energy savings are possible using different precisions while maintaining acceptable accuracy. Further, when operating in low precision/quantization, a portion of the obtained energy benefits can be re-appropriated to recoup the lost accuracy by increasing the network size. As shown in the Figure~\ref{fig:pareto}, this methodology can eliminate the accuracy drop (for example in the case of Power of Two++ (6,16)) while still delivering energy savings of 35.93\%. The figure highlights that larger networks with lower precision can dominate the full-precision baseline design in both accuracy and energy requirements.

